\documentclass{article}
\usepackage{spconf,amsmath}
\usepackage[dvips]{graphicx}
\usepackage{graphicx}
\usepackage{amssymb}
\usepackage{cite}
\usepackage{subfigure}
\usepackage{algorithm}
\usepackage{algorithmic}
\usepackage{multirow}

\title{Simultaneous Block-Sparse Signal Recovery Using Pattern-Coupled Sparse Bayesian Learning}
\name{Hang Xiao$^1$, Zhengli Xing $^2$, Linxiao Yang$^1$, Jun Fang$^1$, and Yanlun Wu$^1$}
\address{$^1$National
	Key Laboratory of Science and Technology on Communications,\\
	University of Electronic Science and Technology of China,
	Chengdu 611731, China,\\
$^2$China Academy of Engineering Physics, Mianyang
		621900, China.}

\begin{document}
%
\maketitle
\begin{abstract}
In this paper, we consider the block-sparse signals recovery problem
in the context of multiple measurement vectors (MMV) with
common row sparsity patterns. We develop a new method for recovery of common row sparsity MMV signals,
where a pattern-coupled hierarchical Gaussian prior model is introduced 
to characterize both the block-sparsity of the coefficients and the statistical dependency between neighboring coefficients of the common row sparsity MMV signals.
Unlike many other methods, the proposed method is able to automatically 
capture the block sparse structure of the unknown signal. Our method is developed
using an expectation-maximization (EM) framework. Simulation results show that our proposed method offers
competitive performance in recovering block-sparse common row sparsity
pattern MMV signals.
\end{abstract}
\section{Introduction}
\label{sec:intro}

Compressing sensing is a new paradigm for data acquisition and reconstruction through
exploiting the inherent sparsity signals of interest. In practice, sparse signals
usually have additional structure that can be exploited to enhance the recovery performance. For example, the
atomic decomposition of multi-band signals\cite{MishaliEldar09} or audio
signals\cite{GribonvalBacry03} usually results in block-sparse structure, in which
the non-zeros coefficients occurs in cluster. A number of algorithms, e.g. block-OMP\cite{EldarKuppinger10},
mixed $l_2/l_1$ norm-minimization\cite{EldarMishali10} were proposed to recover
the block-sparse signals. These methods address
only the SMV recovery problem. However, in real world the assumption of signals share the same sparsity pattern hold valid. We usually obtain multiple observations,
where the recovery performance can be enhanced by exploiting the joint estimation.
There are a great deal of signals in the real-world
share the unchanged sparsity pattern over time, such as communication signals are assigned to a specific bands of
frequency spectrums, thus, these communication signals often sharing the same sparsity pattern in frequency domain.
It has been shown that compared to SMV case, the successful recovery rate can be greatly improved using multiple measurement vectors\cite{EldarMishali09}\cite{EldarRauhut10}. Wipf and Rao first introduced SBL to sparse signal
recovery for SMV model, and later extended it to the MMV model, deriving
the MSBL algorithm\cite{WipfNagarajan10}. 
However, MSBL algorithm does not take the block-sparse properties of MMV signals into consideration.
Zhang and Rao develop the BSBL algorithm to solve the temporally correlated block-sparse MMV signals\cite{ZhangRao11}.
Nevertheless, BSBL algorithm requires block-partition known \emph{a priori}.

In this paper, we extend our former work\cite{FangShen15} to the MMV scenario. To exploit the statistical dependencies, we
propose pattern-coupled hierarchical Gaussian prior model which characterizes both the block sparseness of the coefficient and the statistical
dependency between neighboring coefficients. A key assumption in the considered MMV model is that the support ( i.e. the indexes of nonzero entries)
of each column in MMV signals are identical\cite{MalioutovCetin05}, and the
block-structure of each signal is entirely unkown\cite{ZhangRao11}. In our hierarchical Bayesian model, the prior for each
coefficient not only involves its own hyperparameter, but also the hyperparameters of
its immediate neighboring coefficients.  An expectation-maximization (EM)
algorithm is developed to learn and the hyperparameters and to estimate the block-sparse MMV signals.
Simulation results are provided to illustrate the effectiveness of the proposed algorithm.

\section{Problem Formulation}

We consider the problem of simultaneously recovering a set of block-sparse signals
$\boldsymbol{x}_l\in\mathbb{R}^{N\times 1}, \forall l=1\dots L$ from an basic
underdetermined system
\begin{align}
	\boldsymbol{y}_l=\boldsymbol{\Phi}\boldsymbol{x}_l+\boldsymbol{v}_l,\qquad\forall l=1\dots L,\label{equ-1}
\end{align}
where $\boldsymbol{y}_l\in\mathbb{R}^{M\times 1}$,
$\boldsymbol{\Phi}\in\mathbb{R}^{M\times N}(M<N)$ and
$\boldsymbol{v}_l\in\mathbb{R}^{M\times 1}$ denote the $l$th measurements,
the sensing matrix and the noise, respectively.
The signal $\boldsymbol{x}_l$ has a block-sparse structure
and all $\{\boldsymbol{x}_l\}$ share the same support. We note
that the block partition of $\boldsymbol{x}_l$ is unknown.
We aim to recover $\{\boldsymbol{x}_l\}$ by exploiting their block-sparsity
and the property that sharing the same row sparsity pattern.

It is easy to see that the model (\ref{equ-1}) can be rewritten in matrix form, given by
\begin{align}
	\boldsymbol{Y}=\boldsymbol{\Phi}\boldsymbol{X}+\boldsymbol{V},
\end{align}
where
$\boldsymbol{Y}\triangleq[\boldsymbol{y}_1,\ldots,\boldsymbol{y}_L]\in\mathbb{R}^{M\times L}$,
$\boldsymbol{X}\triangleq[\boldsymbol{x}_1,\ldots,\boldsymbol{x}_L]\in\mathbb{R}^{N\times L}$,
and $\boldsymbol{V}\triangleq[\boldsymbol{v}_1,\ldots,\boldsymbol{v}_L]\in\mathbb{R}^{M\times L}$
is unknown noise matrix. We assume that the elements of $\boldsymbol{V}$ are i.i.d white
noise following Gaussian distribution with zero mean and $\lambda^{-1}$ variance.
Then the distribution of $\boldsymbol{Y}$ conditional on $\boldsymbol{X}$ is
given as
\begin{align}
	p(\boldsymbol{Y}|\boldsymbol{X})=\left(\frac{\lambda}{2\pi}\right)^{\frac{ML}{2}}\exp\left(-\frac{\lambda}{2}\|\boldsymbol{Y}-\boldsymbol{\Phi}\boldsymbol{X}\|_F^2\right)
\end{align}


To simultaneously capture the property
of column-wise block-sparsity and the common row sparsity pattern of $\boldsymbol{X}$,
we assign a Gaussian prior on each row of $\boldsymbol{X}$. Specifically, we impose
a Gaussian prior distribution on the $n$th row of $\boldsymbol{X}$, i.e. $\boldsymbol{x}_{n\cdot}$, with zero mean
and $(\alpha_n+\beta\alpha_{n-1}+\beta\alpha_{n+1})^{-1}\boldsymbol{B}_2^{-1}$ covariance matrix, i.e.,
\begin{align}
p(\boldsymbol{x}_{n\cdot})=\mathcal{N}(\boldsymbol{0},(\alpha_n+\beta\alpha_{n-1}+\beta\alpha_{n+1})^{-1}\boldsymbol{B}_2^{-1})\label{equ-2}
\end{align}
where $\boldsymbol{B}_2^{-1}$ is a positive matrix characterizing the dependency
of the elements of $\boldsymbol{x}_{n\cdot}$. We note that all the
rows of $\boldsymbol{X}$ share the same $\boldsymbol{B}_2$ which has been shown
that such a prior is able to promote the low-rankness
of $\boldsymbol{X}^T$\cite{XinWang2016}, i.e.,
automatically capture the correlation among the rows of $\boldsymbol{X}$.
In (\ref{equ-2}), $\{\alpha_n\}_{n=0}^N$ are positive scalars controlling the sparsity of
the rows of $\boldsymbol{X}$. We assume $\alpha_0=0$ and $\alpha_{N+1}=0$ for 
the end rows $\boldsymbol{x}_{1}$
and $\boldsymbol{x}_{n}$, $0 \le\beta \le1$ is a parameter indicating the relevance
between the coefficient $\boldsymbol{x}_{n}$ and its neighboring coefficients$\{\boldsymbol{x}_{{n+1}},\boldsymbol{x}_{{n-1}}\}$.
It has been shown
that by coupling the neighbor elements using $\{\alpha_n\}$, such a prior has potential to encourage a block-sparse solution\cite{FangShen15}. Then the joint distribution
of $\boldsymbol{X}$ given $\{\alpha_n\}$ and $\boldsymbol{B}_2$ can be written
as
\begin{align}
	p(\boldsymbol{X})=\frac{|\boldsymbol{B}_1|^{\frac{L}{2}}|\boldsymbol{B}_2|^{\frac{N}{2}}}{\sqrt{(2\pi)^{NL}}}\exp\left(-\frac{\text{tr}(\boldsymbol{X}^T\boldsymbol{B}_1\boldsymbol{X}\boldsymbol{B}_2)}{2}\right).
\end{align}
where
$\boldsymbol{B}_1$ to be a diagonal matrix with its $n$th diagonal element equals
to $(\alpha_n+\beta\alpha_{n-1}+\beta\alpha_{n+1})$.

\section{Proposed Bayseian Inference Algorithm}

In this section, we proceed to develop a sparse Bayesian learning
method for block-sparse MMV signal recovery.
Based on the above
hierarchical model, the posterior distribution of $\boldsymbol{X}$
can be computed as
\begin{align}
	p(\boldsymbol{X}|\boldsymbol{Y};\boldsymbol{\alpha},\boldsymbol{B}_2,\lambda)\propto p(\boldsymbol{Y}|\boldsymbol{X};\lambda)p(\boldsymbol{X};\boldsymbol{\alpha},\boldsymbol{B}_2)
\end{align}
where $\boldsymbol{\alpha}=\{\alpha_n\}_{n=1}^N$.

The log-posterior of $\boldsymbol{X}$ can be written as
\begin{align}
	&\ln p(\boldsymbol{X}|\boldsymbol{Y};\boldsymbol{\alpha},\boldsymbol{B}_2,\lambda)\nonumber\\
	\propto&-\frac{\lambda}{2}\|\boldsymbol{Y}-\boldsymbol{\Phi}\boldsymbol{X}\|_F^2-\frac{1}{2}\text{tr}(\boldsymbol{X}^T\boldsymbol{B}_1\boldsymbol{X}\boldsymbol{B}_2)\nonumber\\
	\propto&-\frac{1}{2}\boldsymbol{x}^T(\lambda\boldsymbol{I}\otimes(\boldsymbol{\Phi}^T\boldsymbol{\Phi})+\boldsymbol{B}_2\otimes\boldsymbol{B}_1)\boldsymbol{x}-\lambda\boldsymbol{y}^T(\boldsymbol{I}\otimes\boldsymbol{\Phi})\boldsymbol{x}\nonumber
\end{align}
where $\boldsymbol{x}$ and $\boldsymbol{y}$ denote the vectorization of $\boldsymbol{X}$
and $\boldsymbol{Y}$, respectively, and $\otimes$ denotes the operation of
Kronecker product.
Then we arrive at that the posterior distribution of $\boldsymbol{x}$
follows the Gaussian distribution with mean and covariance matrix given as
\begin{align}
	\boldsymbol{\mu}&=\lambda\boldsymbol{\Sigma}(\boldsymbol{I}\otimes\boldsymbol{\Phi})^T\boldsymbol{y}\label{x-mean}\\
	\boldsymbol{\Sigma}&=(\lambda\boldsymbol{I}\otimes(\boldsymbol{\Phi}^T\boldsymbol{\Phi})+\boldsymbol{B}_2\otimes\boldsymbol{B}_1)^{-1} \label{x-var}
\end{align}

Given a set of estimated hyperparameters $\boldsymbol{\alpha}$, $\boldsymbol{B}_2$, $\lambda$ and the observed $\boldsymbol{y}$, the maximum a posterior (MAP)
estimate of $\boldsymbol{x}$ is the mean of its posterior distribution, i.e.
\begin{align}
	\hat{\boldsymbol{x}}_{\mathrm{MAP}}=\boldsymbol{\mu}
\end{align}

Our problem therefore reduces to estimate the value of the
hyperparameters $\boldsymbol{\alpha}$, $\boldsymbol{B}_2$, and $\lambda$.
A strategy to maximize the likelihood function of these hyperparameters
is to exploit the expectation-maximization (EM) formulation,
in which we first introduce a hidden variable and then iteratively
maximize a lower bound of the likelihood function (this lower bound
is also referred to as the Q-function). Briefly speaking, the
algorithm alternates between an E-step and a M-step. In the
E-step, we compute a new Q-function by taking
expectation of the log joint distribution of data and hidden variable
with respect to the posterior of hidden variable which computed using current estimation
of the hyperparameters. In the M-step, we update the hyperparameters
by maximizing the Q-function with respect to them.

We define the hyperparameters
$\boldsymbol{\Theta}=\{\boldsymbol{\alpha},\boldsymbol{B}_2,\lambda\}$
and recognize $\boldsymbol{X}$ as the hidden variable. Then
Q-function can be expressed as
\begin{align}
	{Q}(\boldsymbol{\Theta})&=E_{p(\boldsymbol{X}|\boldsymbol{Y};\boldsymbol{\Theta}^{(t)})}[\ln p(\boldsymbol{Y},\boldsymbol{X};\boldsymbol{\Theta})],
\end{align}
and, consequently, $\boldsymbol{\Theta}$ can be updated by maximizing $Q$-function,
i.e.,
\begin{align}
\boldsymbol{\alpha}^{(t+1)}&=\arg\max_{\boldsymbol{\alpha}} Q(\boldsymbol{\Theta})\nonumber\\
&=\arg\max_{\boldsymbol{\alpha}} E_{p(\boldsymbol{X};\boldsymbol{Y},\boldsymbol{\Theta}^{(t)})}[\ln p(\boldsymbol{X};\boldsymbol{\alpha}_1,\boldsymbol{B}_2)]\label{Q-alpha}\\
\boldsymbol{B}_2^{(t+1)}&=\arg\max_{\boldsymbol{B}_2}
Q(\boldsymbol{\Theta})\nonumber\\
&=\arg\max_{\boldsymbol{B}_2} E_{p(\boldsymbol{X}|\boldsymbol{Y};\boldsymbol{\Theta}^{(t)})}[\ln p(\boldsymbol{X};\boldsymbol{\alpha}_1,\boldsymbol{B}_2)]\label{Q-B2}\\
\lambda^{(t+1)}&=\arg\max_{\lambda}
Q(\boldsymbol{\Theta})\nonumber\\
&=\arg\max_{\lambda} E_{p(\boldsymbol{X}|\boldsymbol{Y};\boldsymbol{\Theta}^{(t)})}[\ln p(\boldsymbol{Y},\boldsymbol{X};\lambda)]\label{Q-lambda}
\end{align}
%

%
%

We first evaluate the expectation in (\ref{Q-alpha}) and (\ref{Q-B2}), which is given as
\begin{align}
&E_{p(\boldsymbol{X}|\boldsymbol{Y};\boldsymbol{\Theta}^{(t)})}[\ln p(\boldsymbol{X}|\boldsymbol{\alpha}_1,\boldsymbol{B}_2)]\nonumber\\
=&\Big\langle\frac{L}{2}\ln|\boldsymbol{B}_1|+\frac{N}{2}\ln|\boldsymbol{B}_2|
-\frac{\text{tr}(\boldsymbol{X}^T\boldsymbol{B}_1\boldsymbol{X}\boldsymbol{B}_2)}{2}\Big\rangle\nonumber\\
=&\frac{L}{2}\ln|\boldsymbol{B}_1|+\frac{N}{2}\ln|\boldsymbol{B}_2|
-\frac{1}{2}\text{tr}((\boldsymbol{B}_2\otimes\boldsymbol{B}_1)\langle\boldsymbol{x}\boldsymbol{x}^T\rangle)\label{Q-exp}
\end{align}
where $\langle\cdot\rangle$ denotes the operator that taking expectation using
the distribution $p(\boldsymbol{X}|\boldsymbol{Y};\boldsymbol{\Theta}^{(t)})$,
and $\langle\boldsymbol{x}\boldsymbol{x}^T\rangle$ is given as
\begin{align}
\langle\boldsymbol{x}\boldsymbol{x}^T\rangle=\boldsymbol{\mu}\boldsymbol{\mu}^T+\boldsymbol{\Sigma}
\end{align}

Then the problem (\ref{Q-alpha}) can be solved by setting the first
derivative of the (\ref{Q-exp}) with respect to $\alpha_1$ to zero, i.e.,
\begin{align}
\frac{\partial {Q}(\boldsymbol{\Theta})}{\partial \alpha_i}=
\frac{L}{2}(\nu_i+\beta\nu_{i-1}+\beta\nu_{i+1})-\phi_i
=0\label{alpha-prob}
\end{align}
where we define $\nu_i\triangleq(\alpha_i+\beta\alpha_{i+1}+\beta\alpha_{i-1})^{-1},
\forall i=1,\ldots,M$, with
$\nu_0=\nu_{M+1}=0$, and $\phi_i\triangleq\frac{1}{2}(\text{Tr}(\boldsymbol{B}_2\boldsymbol{\Omega}_i)
+\beta\text{Tr}(\boldsymbol{B}_2\boldsymbol{\Omega}_{i-1})+\beta\text{Tr}(\boldsymbol{B}_2\boldsymbol{\Omega}_{i+1}))$,
in which $\boldsymbol{\Omega}_i\triangleq\langle\boldsymbol{x}_{i\cdot}\boldsymbol{x}_{i\cdot}^T\rangle$,
with $\boldsymbol{x}_{i\cdot}$ denotes the $i$the row of $\boldsymbol{X}$.
Similarly, we set $\boldsymbol{\Omega}_0=\boldsymbol{\Omega}_{N+1}=\boldsymbol{0}$.
Then the optimal solution $\alpha_i^{*}$ should satisfy
\begin{align}
\frac{L}{2}(\nu_i^*+\beta\nu_{i-1}^*+\beta\nu_{i+1}^*)=\phi_i
\end{align}
Since all the hyper parameters$\{  \alpha_i  \}$ are non-negative, we have
\begin{align}
\frac{1}{\alpha_i^*}>\nu_i^*>0, \forall i=1,\ldots,N
\\
\frac{1}{\beta\alpha_{i+1}^*}>\nu_i^*>0, \forall i=1,\ldots,N-1
\\
\frac{1}{\beta\alpha_{i-1}^*}>\nu_i^*>0, \forall i=2,\ldots,N
\end{align}
\par
Hence the term on the left-hand side of (27) is lower and upper bounded
respectively by
\begin{align}
\frac{3L}{2\alpha_i^*}>\phi_i>0
\end{align}
combining above equations (27)-(31), we arrive at
\begin{align}
\alpha_i^*\in\big[0,\frac{3L}{2\phi_i}\big]
\end{align}
Due to the high computational complexity of calculating an accuracy solution
of (\ref{alpha-prob}), we employ an sub-optimal solution of it, i.e., just set
$\alpha_i$ to its upper bound, which arrives at
\begin{align}
\alpha_i^{(t+1)}=\frac{3L}{2\phi_i}\label{Update-alpha}
\end{align}
Although We employ a sub-optimal solution (\ref{Update-alpha}) to update the hyperparameter in M-step, numerical results show that the sub-optimal update rule is quite effective. This is because the sub-optimal solution (\ref{Update-alpha}) provide a reasonable estimate of the optimal solution.

We then consider solving the problem (\ref{Q-B2}). Similarly, we
set the first derivative of the Q-function with respect to $\boldsymbol{B}_2$ to zero,
i.e.,
\begin{align}
\frac{\partial {Q}(\boldsymbol{\Theta})}{\partial \boldsymbol{B}_2}=
\frac{N}{2}\boldsymbol{B}_2^{-1}-\frac{1}{2}&\sum\limits_{i=1}^{N}\nu_i^{-1}\boldsymbol{\Omega}_i
\end{align}
and arrive at that the optimal solution is given by
\begin{align}
\boldsymbol{B}^{(t+1)}=\frac{1}{N}&\sum\limits_{i=1}^{N}(\alpha_i+\beta\alpha_{i-1}+\beta\alpha_{i+1})^{-1}\boldsymbol{\Omega}_i^{-1}\label{Update-B}
\end{align}

To estimate $\lambda$, the Q-function can be simplified to
\begin{align}
{Q}(\lambda)=& E_{{x}|{y},\Theta^{(t)}}[\log p(\boldsymbol{y}|\boldsymbol{x};\lambda)]
\nonumber \\
\propto &\frac{NL}{2}\log \lambda-\frac{\lambda}{2}\langle\| \boldsymbol{Y}-\boldsymbol{\Phi}\boldsymbol{X}\|_F^2\rangle\label{Q-lambda1}
\end{align}

By computing the derivative of (\ref{Q-lambda1}) and setting it to zero,
we arrive at
\begin{align}
\lambda^{(t+1)}&=\frac{1}{NL}\langle\|\boldsymbol{Y}-\boldsymbol{\Phi}\boldsymbol{X}\|_F^2\rangle\nonumber\\
&=\frac{1}{NL}\langle\|\boldsymbol{y}-(\boldsymbol{I}\otimes\boldsymbol{\Phi})\boldsymbol{x}\|_2^2\rangle\label{Update-lambda}
\end{align}

Some of the expectations and
moments used during the update are summarized as
\begin{align}
&\langle\boldsymbol{x}\boldsymbol{x}^T\rangle=\boldsymbol{\mu}\boldsymbol{\mu}^T+\boldsymbol{\Sigma}\qquad
\langle\boldsymbol{x}_{i\cdot}\boldsymbol{x}_{i\cdot}^T\rangle=\boldsymbol{\mu}_i\boldsymbol{\mu}_i^T+\boldsymbol{\Sigma}_i\\
&\langle\|\boldsymbol{y}-\boldsymbol{A}\boldsymbol{x}\|_2^2\rangle=\|\boldsymbol{y}-\boldsymbol{A}\boldsymbol{\mu}\|_2^2+\text{tr}(\boldsymbol{A}^T\boldsymbol{A}\boldsymbol{\Sigma})
\end{align}
where $\boldsymbol{\mu}_i$ is a vector with its $l$th element equals to
the $((i-1)L+l)$th of $\boldsymbol{\mu}$'s and $\boldsymbol{\Sigma}_i$ is a matrix
with its $(n_1,n_2)$th entry equals to the $((i-1)L+n_1,(i-1)L+n_2)$th of $\boldsymbol{\Sigma}$.

For clarity, we summarize our algorithm as follows.
\begin{algorithm}
	\renewcommand{\algorithmicrequire}{\textbf{Input:}}
	\renewcommand\algorithmicensure {\textbf{Output:}}
	\caption{Pattern-coupled sparse Bayesian learning algorithm for multiple measurement vector}
	\begin{algorithmic}[1]
		\REQUIRE $\boldsymbol{y}$, $\boldsymbol{\Phi}$ and $\beta$
		\ENSURE $\boldsymbol{x}$, $\boldsymbol{\alpha}$, $\boldsymbol{B}_2$, $\gamma$
		\STATE Select an initialization $\boldsymbol{x}^{(0)}$, $\boldsymbol{\alpha}^{(0)}$, $\boldsymbol{B}_2^{(0)}$, $\gamma^{(0)}$, and set $t=0$.
		\WHILE {not converged}
		\STATE Calculate $\boldsymbol{\alpha}^{(t+1)}$, $\boldsymbol{B}_2^{(t+1)}$ and $\gamma^{(t+1)}$ according to (\ref{Update-alpha}), (\ref{Update-B}), and (\ref{Update-lambda}), respectively.
		\STATE According to (\ref{x-mean}) and (\ref{x-var}), update the posterior distribution of $\boldsymbol{X}$ using $\boldsymbol{\alpha}^{(t+1)}$, $\boldsymbol{B}_2^{(t+1)}$ and $\gamma^{(t+1)}$.
		\STATE $t=t+1$
		\ENDWHILE
	\end{algorithmic}
	\label{algorithm}
\end{algorithm}


\section{Simulation Results}

In our simulations, we study how the proposed algorithm benefit from multiple measurement vectors for
the block-sparse common row sparsity pattern signals recovery problem. Suppose each $N$-dimensional sparse vector contains
$K$ nonzero coefficients which are partitioned into blocks with random sizes and random location. The over-complete $M \times N$ dictionary $\boldsymbol{\Phi}$ are
randomly generated with each entry independently drawn from a normal distribution.
\par

\begin{figure}[t]
	\centering
	\includegraphics [width=8cm,height=6cm]{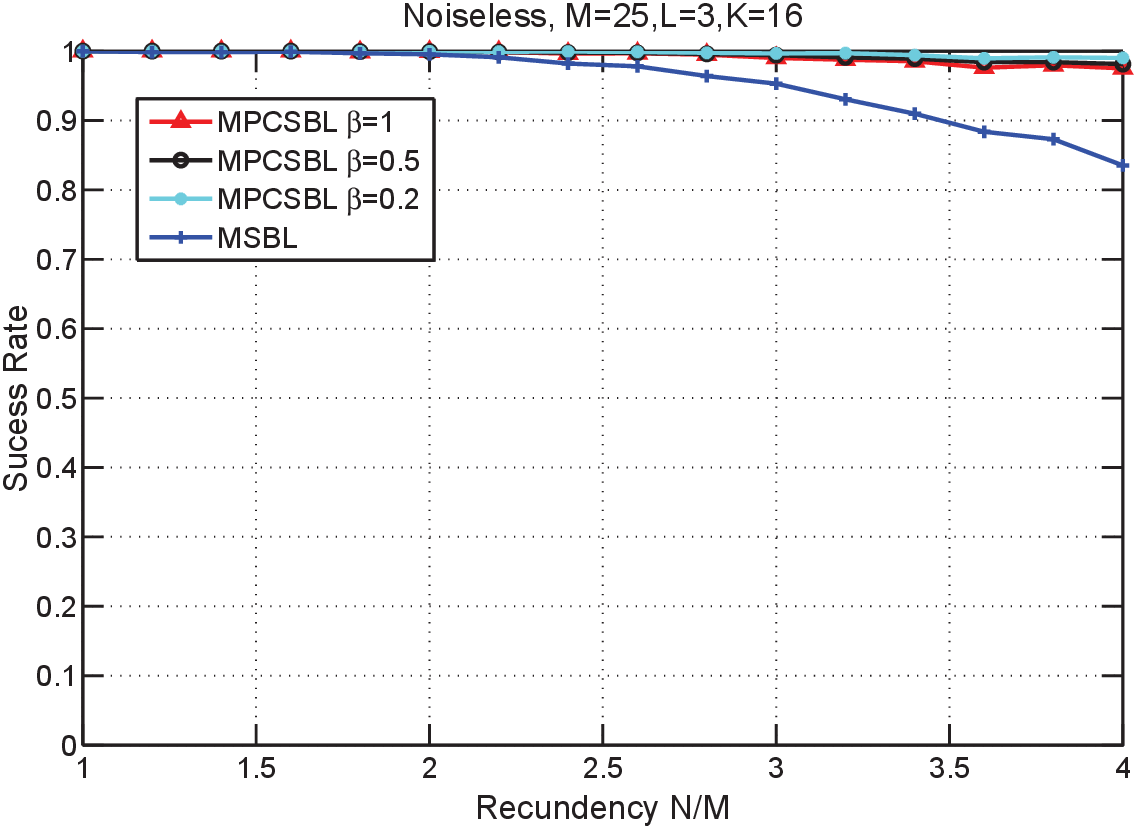}\\
	\caption{Success Rate vs. N/M}
	\label{fig:RatioMN}
\end{figure}
\begin{figure}[t]
	\centering
	\includegraphics [width=8cm,height=6cm]{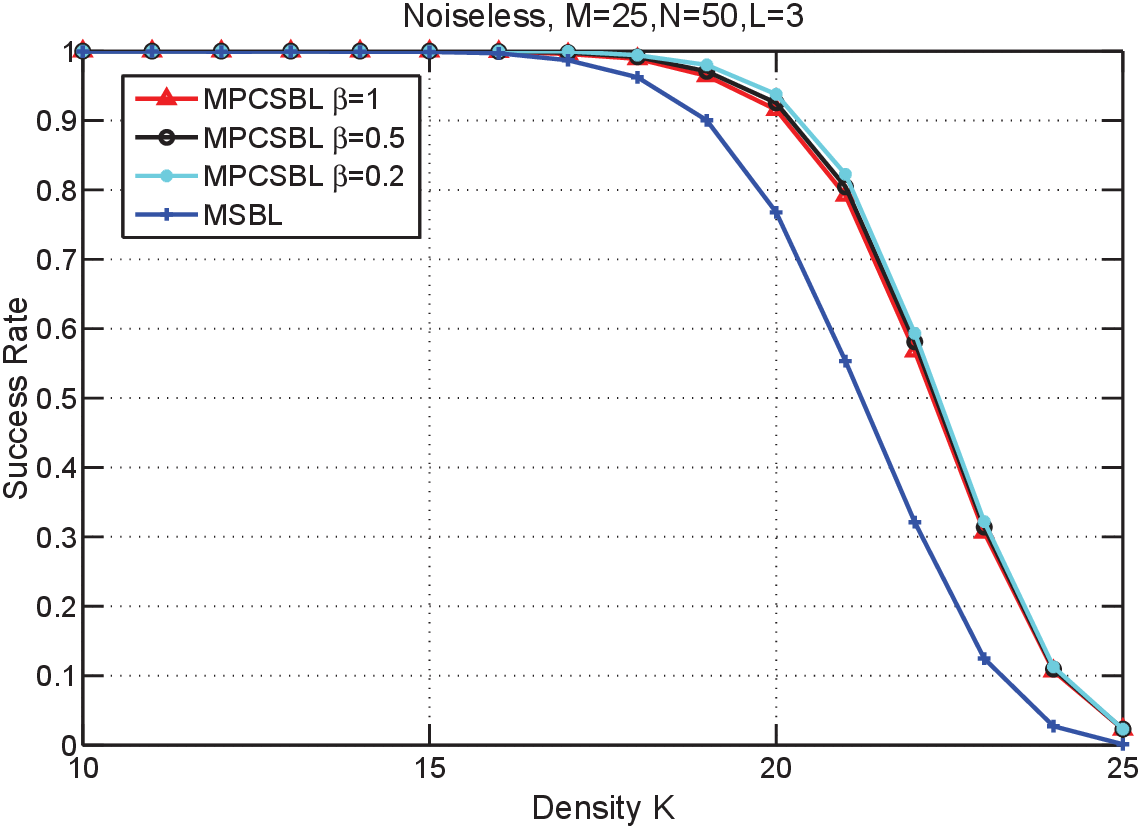}\\
	\caption{Success Rate vs. Nonzero Source Number K}
	\label{fig:density}
\end{figure}
\begin{figure}[t]
	\centering
	\includegraphics [width=8cm,height=6cm]{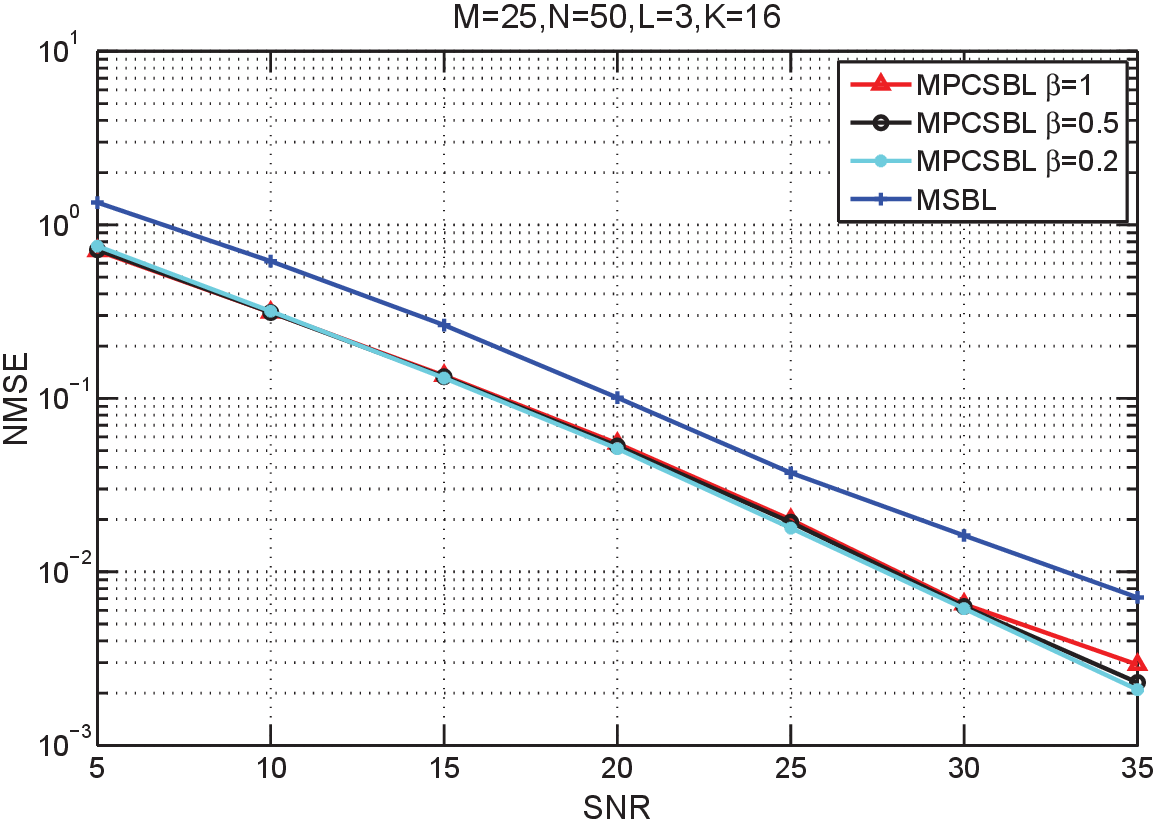}\\
	\caption{NMSE vs. SNR}
	\label{fig:SNR_MSE}
\end{figure}
We examine the recovery performance of our proposed algorithm, also referred as the MMV pattern-coupled sparse Bayesian learning algorithm (MPCSBL), under different
choice of $\beta$. As indicated earlier in our paper, $\beta,(0\leq \beta \leq 1) $ is a parameter quantifying the dependencies among the neighboring coefficients.
Fig.\ref{fig:RatioMN} depicts the success rates vs. the ratio $N/M$ for different choices of $\beta$ in noiseless case, where we set $M=25$, $L=3$, $K=16$.
Results are averaged over 1000 independent runs, with the measurement matrix and the sparse signal randomly generated for each run.
The performance of conventional sparse Bayesian learning method ( donated as "MSBL"\cite{ZhangRao11}) is also included for our comparision.
When $\beta>0$, our proposed algorithm achieves a significant performance improvement as compared with MSBL through
exploiting the underlying block-sparse structure, even without knowing the exacting locations and the sizes of the
non-zero blocks. We also observe that our proposed algorithm is not very sensitive to the choice of $\beta$ as long
as $\beta>0$. The success rates of proposed method and MSBL 
as a function of the sparsity level are plotted in Fig.\ref{fig:density}, 
where $M=25$, $N=50$ and $L=3$. We see that our proposed algorithm present the better performance than the MSBL method. In noisy case, we setting $M=25$, $N=50$, $K=16$ and $L=3$, Fig.\ref{fig:SNR_MSE} shows that the normalized mean square errors (NMSE) vs. SNR  with $K$ nonzero rows for different choices of $\beta$. We also see that our proposed method achieves the better estimation accuracy than existing methods.

\section{Conclusion}
\label{sec:majhead}

We proposed a new Bayesian method for recovery block-sparse MMV signals with common row sparsity pattern. A pattern-coupled hierarchical Gaussian prior model
was introduced to characterize both the sparseness of the coefficients and the statistical
dependency between
neighboring coefficients of the signal. Through exploiting the underlying block-structure, our method outperforms other existing methods in block-sparse MMV signals recovery with common row sparsity pattern. Numerical results show that the
proposed method presents superior performance in recovery block-sparse MMV signals with common row sparsity pattern.

\vfill\pagebreak

\bibliography{newbib}
\bibliographystyle{IEEEbib}

\end{document}